\documentclass{article}

\usepackage[preprint]{neurips_2022}
\usepackage{graphicx}

\usepackage[utf8]{inputenc} 
\usepackage[T1]{fontenc}    
\usepackage{hyperref}       
\usepackage{url}            
\usepackage{booktabs}       
\usepackage{amsfonts}       
\usepackage{nicefrac}       
\usepackage{microtype}      
\usepackage{xcolor}         
\usepackage{caption}

\title{Testing Relational Understanding in Text-Guided Image Generation}

\author{%
  Colin Conwell \\
  Department of Psychology\\
  Harvard University\\
  Cambridge, MA, 02138 \\
  \texttt{conwell@g.harvard.edu} \\
  \And
  Tomer D. Ullman \\
  Department of Psychology \\
  Harvard University \\
  Cambridge, MA, 02138 \\
  \texttt{tullman@fas.harvard.edu} \\
}

\begin{document}

\maketitle

\begin{abstract}
Relations are basic building blocks of human cognition. Classic and recent work suggests that many relations are early developing, and quickly perceived. Machine models that aspire to human-level perception and reasoning should reflect the ability to recognize and reason generatively about relations. We report a systematic empirical examination of a recent text-guided image generation model (DALL-E 2), using a set of 15 basic physical and social relations studied or proposed in the literature, and judgements from human participants (N = 169). Overall, we find that only $\sim$22\% of images matched basic relation prompts. Based on a quantitative examination of people's judgments, we suggest that current image generation models do not yet have a grasp of even basic relations involving simple objects and agents. We examine reasons for model successes and failures, and suggest possible improvements based on computations observed in biological intelligence. 

\end{abstract}

\section{Introduction}

Consider the line `the flooben was on the demaglis'. Even if you don't know what a \textit{flooben} or \textit{demaglis} are, you know something is \textit{on} something\footnote{As Alice remarks after reading the nonsense poem, Jabberwockey: \textit{``Somehow it seems to fill my head with ideas -- only I don’t exactly know what they are! However, somebody killed something: that’s clear, at any rate''}}. This is because \textit{on} is a basic relation. Our understanding of basic relations is general, early developing \citep{hespos2004conceptual}, and fundamental to our reasoning \citep{talmy1985lexicalization}. There is also growing evidence that basic relations are perceived as directly as basic object properties \citep{hafri2021perception}. Machines that attempt to capture elements of human reasoning would do well to accurately perceive such relations in images, and produce accurate images from such relations as input. Here, we empirically and systematically evaluate a recent state-of-the-art model for image generation (DALL-E 2) on its understanding of basic relations. 

Recent advances in image synthesis have achieved seemingly remarkable success in producing arbitrary images from arbitrary text \citep[e.g.][]{saharia2022photorealistic, ramesh2022hierarchical}. A prompt such as `a robot-cat wearing cool glasses, gazing at a supernova' produces images that look somewhat like a robot-cat, wearing cool glasses, gazing at a supernova. Such successes lead to the impression that these models understand the input as a human would, as a compositional combination of objects, properties, and relations. 

Despite their success, these models are not without their limitations. Marcus, Davis, and Aaronson used an informal preliminary analysis to illustrate the limitations of DALL-E 2 in compositionality, common sense, anaphora, relations, negation, and number \cite{marcus2022very}. AI blogger `Swimmer963' \cite{swimmer963} reported informal tests along similar lines, and concluded DALL-E 2 has weaknesses with multiple characters, text, novel words, and foreground-background. Farid \cite{farid2022perspective} has pointed out the implausibility of cast shadows and reflections in DALL-E 2. Liu et al. \citep{liu2022compositional} recently proposed a composable diffusion model, and show that it outperforms other text-to-image models in the generation of structured images, by using basic conjunction (AND) and negation (NOT). 

Some of the limitations of current image-generation models have been recognized  by the developers of the models themselves. For example, Ramesh et al. point out difficulties with binding, relative size, text, and other issues \citep[Section 7 in][] {ramesh2022hierarchical}. Saharia et al. proposed the DrawBench benchmark, which includes a head-to-head comparison of the Imagen model to DALL-E 2, GLIDE \citep{nichol2021glide}, VQ-GAN-CLIP \cite{crowson2022vqgan}, and Latent Diffusion \cite{rombach2022high}, on images that probe the limitations pointed out in prior work, including number, unorthodox color, positional arguments, rare words, and text generation. 

While informative and important, these tests have not yet focused systematically on basic relations, and have been restricted to non-relational limitations, a small number of prompts, a head-to-head comparison of models to models rather than models to people, the intuitions of the authors, long and complex prompts, or some combination of all of these factors. 

The current work focuses on a set of 15 basic relations previously described, examined, or proposed in the cognitive, developmental, or linguistic literature. The set contains both grounded spatial relations (e.g. 'X on Y'), and more abstract agentic relations (e.g. 'X helping Y'). The prompts are intentionally simple, without attribute complexity or elaboration. That is, instead of a prompt like `a donkey and an octopus are playing a game. The donkey is holding a rope on one end, the octopus is holding onto the other. The donkey holds the rope in its mouth. A cat is jumping over the rope', we use `a box on a knife'. The simplicity still captures a broad range of relations from across various subdomains of human psychology, and makes potential model failures more striking and specific. 

Rather than rely on our own intuition for whether an image matches a given relation prompt, we examined the intuitions of 169 participants. The use of multiple relations and many participants allows a more nuanced examination of model performance than pass/fail judgements. It also allows a quantitative examination of model performance when considering additional covariates, such as the ranking of images by CLIP score. The stimulus set, prompts, images, and participant data are all openly available at \url{https://osf.io/sm68h}. 

\section{Background}

The majority of text-guided image generation algorithms are a combination of two machine learning techniques: reconstructive generative modeling, and latent space manipulation (steering) by way of natural language supervision. DALL-E 2 in particular uses a combination of latent diffusion modeling and CLIP-style natural language supervision \citep{ramesh2022hierarchical}. Latent diffusion modeling is a reconstructive generative modeling technique that builds hierarchical representations of data by taking inputs, introducing noise, then teaching the model to reconstruct the original (noiseless) data through progressive transformation \citep{rombach2022high}. This technique allows the model to learn a structured, low-dimensional prior over image space that can serve as the basis for the generation of entirely novel, high-dimensional images by way of (re)sampling. 

CLIP is a method for linking image-to-text by way of two encoders that reduce a paired image-text sample to equidimensional latent vectors, and a loss function that forces the similarity of both these vectors to 1 -- effectively tagging both image and text as having originated from the same data-generating source \citep{radford2021learning}. Once trained on a sufficiently diverse set of samples, CLIP can be used to assess the similarity between any image and any text. It is this similarity score that DALL-E 2 (and other algorithms like it) use as a steering signal over the latent space of a trained diffusion model, effectively guiding the diffusion model to produce samples conditioned on the similarity between the generated samples and the target text. While relatively straightforward in its implementation, the success of DALL-E-like models is contingent on the use of massive image-text training sets. The earliest iterations of CLIP, for example, relied on a training set of 400-million image-text pairs, heretofore made unavailable to the wider public. The size of the training set for the most recent iteration of DALL-E 2 used in this experiment has of this writing remained unspecified. 
\vspace{15ex}

\section{Experiment}

We designed our experiment to assess the fit between basic relations and the images formed by DALL-E 2, by presenting images and sentences to human respondents and asking them whether an image and sentence matched. 

Based on the existing cognitive, linguistic, and developmental literature \citep{chen1982topological,lovett2017topological,strickland2011event,kellman1983perception,spelke1994initial,yildirim2016perceiving,gao2009psychophysics,hamlin2007social,ullman2009help,van2016automaticity,glanemann2016rapid,dobel2007describing,guan2020seeing,firestone2017seeing,hafri2020phone}, we created a set of 15 relations (8 physical, 7 agentic). The physical relations were: \textit{in}, \textit{on}, \textit{under}, \textit{covering}, \textit{near}, \textit{occluded by}, \textit{hanging over}, and \textit{tied to}. The agentic relations were: \textit{pushing}, \textit{pulling}, \textit{touching}, \textit{hitting}, \textit{kicking}, \textit{helping}, and \textit{hindering}. These relations were either studied previously in psychophysics, proposed as early developing in humans, proposed as quickly and automatically computed in perception, were the target of computational modeling, are of linguistic interest, or some combination of  these desiderata. 

We created a set of 12 entities: 6 objects, and 6 agents. The objects were: \textit{box}, \textit{cylinder}, \textit{blanket}, \textit{bowl}, \textit{teacup}, and \textit{knife}. The agents were: \textit{man}, \textit{woman}, \textit{child}, \textit{robot}, \textit{monkey}, and \textit{iguana}. The objects were simple bodies or common items used in previous data-sets that study relations \citep[e.g.][]{ehrhardt2020relate,johnson2017clevr}, or in psychophysics tasks \cite{hafri2020phone}, or both. The agents were human, or human-like, or of interest to the AI community. The iguana was a novel visually distinct subordinate category, as a treat. 

\begin{figure*}[b!]
\centering
\vspace{-2.5ex}
\includegraphics[width=0.9\linewidth]{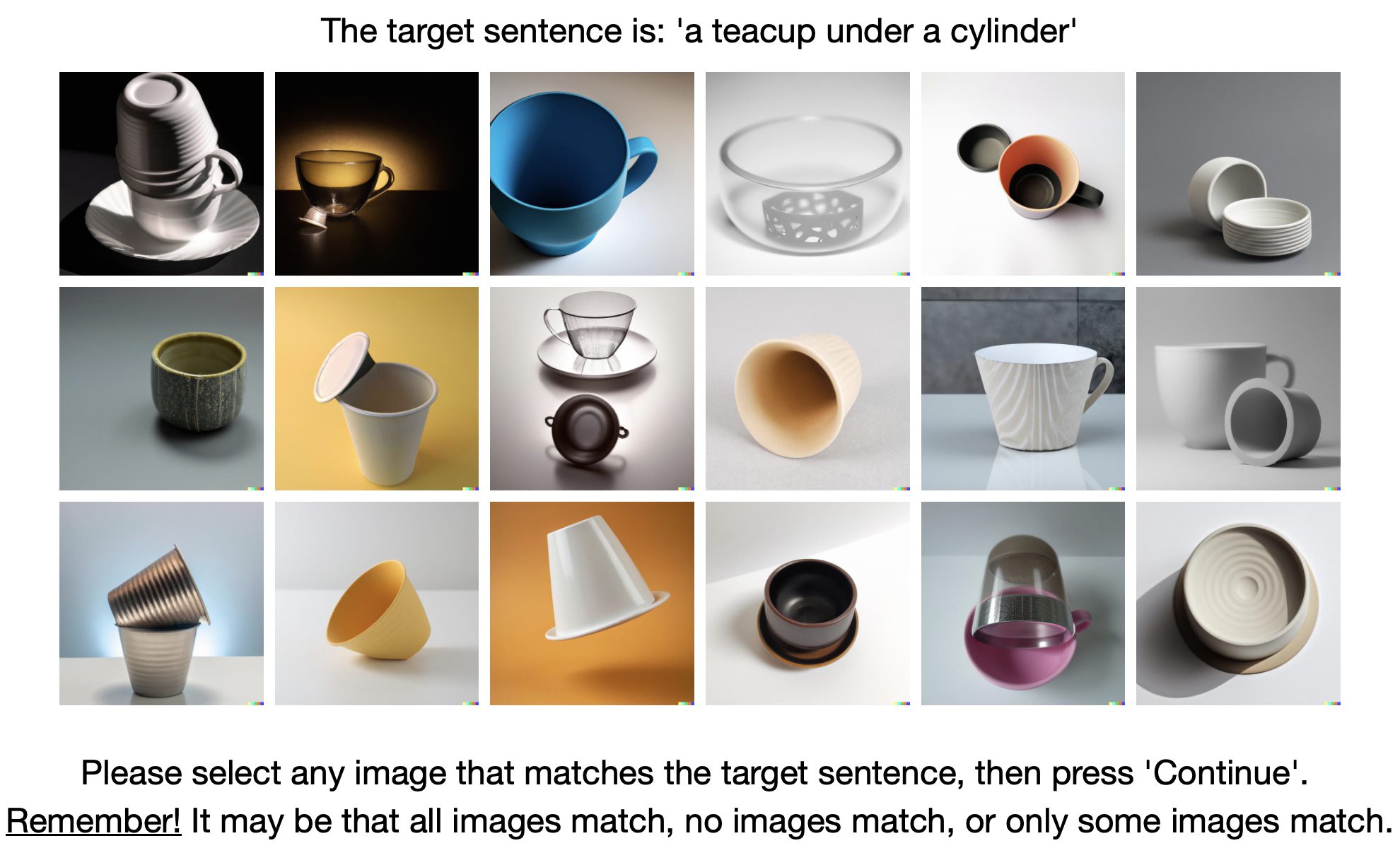}
\captionsetup{justification=justified,margin=0.5cm}
\caption{Screenshot from a trial in our Experiment. Participants were presented with grids of images, and a sentence prompt. Participants selected images that matched the target sentence.}
\label{figure:experiment-screenshot}
\end{figure*}

For each relation, we created 5 different prompts, by randomly sampling two entities five times. This resulted in 75 prompts total (15 relations x 5 samples). For some relations, we restricted the set of allowable entities as follows: (i) Physical relations involved two physical objects, (ii) \textit{Covering} had \textit{blanket} as the first entity, (iii) \textit{In} had \textit{box} or \textit{bowl} as the second entity, (iv) Agentic relations had an agent as the first entity, and either an object or an agent as the second entity, (v) The relations \textit{helping} and \textit{hindering} exclusively involved two agents. 

We submitted each prompt to the DALL-E 2 rendering engine, and obtained the first 18 images that resulted. In a small number of cases, the prompt was rejected as a policy violation (e.g. `a man kicking a man'). In such cases, the second  entity was replaced at random until no policy violation was encountered. Our final stimuli set consisted of 1350 images (75 prompts x 18 images). Prompts, images, and participant data are available at \url{https://osf.io/sm68h}. 

\subsection{Participants} 

We recruited 180 participants online \cite{peer2017beyond} via the Prolific platform (\url{https://www.prolific.co}). Participants were restricted to those located in the USA, having completed at least 100 prior studies on Prolific, with an acceptance rate of at least $90\%$. The mean age of the participants was 33.8; 59\% of participants identified as female, 40\% identified as male, and one did not identify as either. Of this sample, 11 participants failed to pass two attention checks, and were removed from analysis, leaving 169 participants in the final sample. The experiment was approved under an existing IRB (IRB19-1861 Commonsense Reasoning in Physics and Psychology). All participants provided informed consent.

\subsection{Method} 

Participants were informed that they would be assessing a `picture-drawing AI', by examining grids of images that an AI drew in response to a given sentence. After an attention check, participants saw 10 trials, one at a time. In each trial, participants were shown 18 images, organized into a 3x6 grid. Each grid also had the relevant prompt displayed at the top. Participants were instructed to select all images in the grid that match the prompt. Participants were reminded that it may be the case that all images match the prompt, none of them match, or only some of them match (See Figure  \ref{figure:experiment-screenshot}). 

The 10 prompts any given participant rated were randomly drawn from the full set of 75 prompts. This resulted in variability in the number of participants that evaluated any given image. The number of participants that rated a given image ranged from 15 to 43, with an average of 23. After participants finished evaluating 10 prompts, they were given another attention check, thanked for their time, and given an opportunity to provide feedback. 

\subsection{Results} 

Unless otherwise noted, results are reported with the following convention: arithmetic mean [lower 95\% confidence interval, upper 95\% confidence interval].

\begin{figure*}[h!]
\centering
\includegraphics[width=0.8\linewidth]{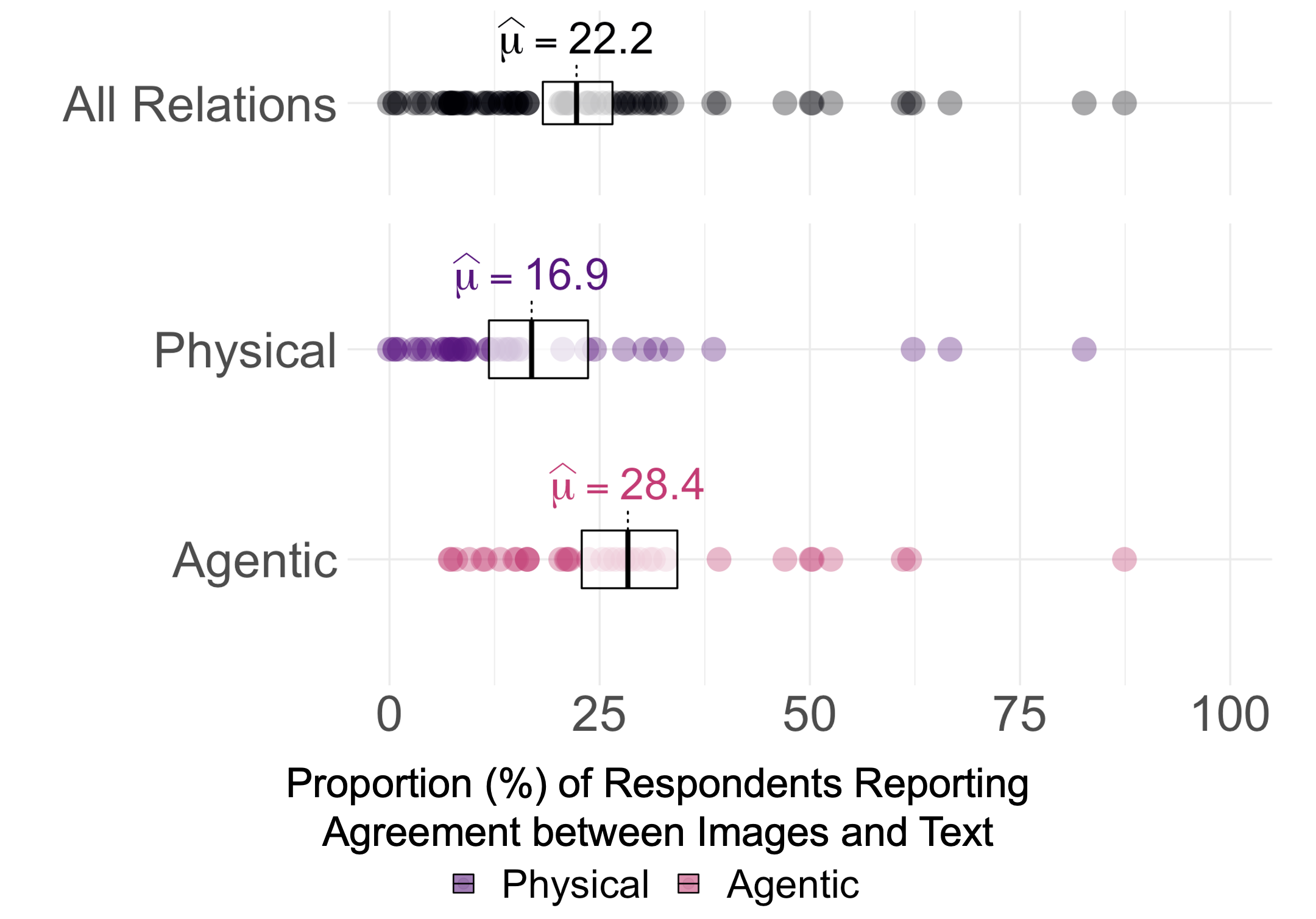}
\captionsetup{justification=justified,margin=0.5cm}
\caption{Experiment results, participant agreement that images matched a prompt. Each point is an individual prompt. Points in black show all prompts. Points in color break down the prompts by whether the subject of the prompt was an object (physical) or agent (agentic).}
\label{figure:overall-scores}
\end{figure*}

Participants on average reported a low amount of agreement between DALL-E 2's images and the prompts used to generate them, with a mean of 22.2\% [18.3, 26.6] across the 75 distinct prompts. Agentic prompts, with a mean of 28.4\% [22.8, 34.2] across 35 prompts, generated higher agreement than physical prompts, with a mean of 16.9\% [11.9, 23.0] across 40 prompts ($t_{Welch}(71.82) = -2.81, p < 8.41e^-3, \hat{g}_{Hedges} = -0.62\:[-1.08,-0.16]$). See also Figure  \ref{figure:overall-scores}.

\begin{figure*}[h!]
\centering
\includegraphics[width=0.85\linewidth]{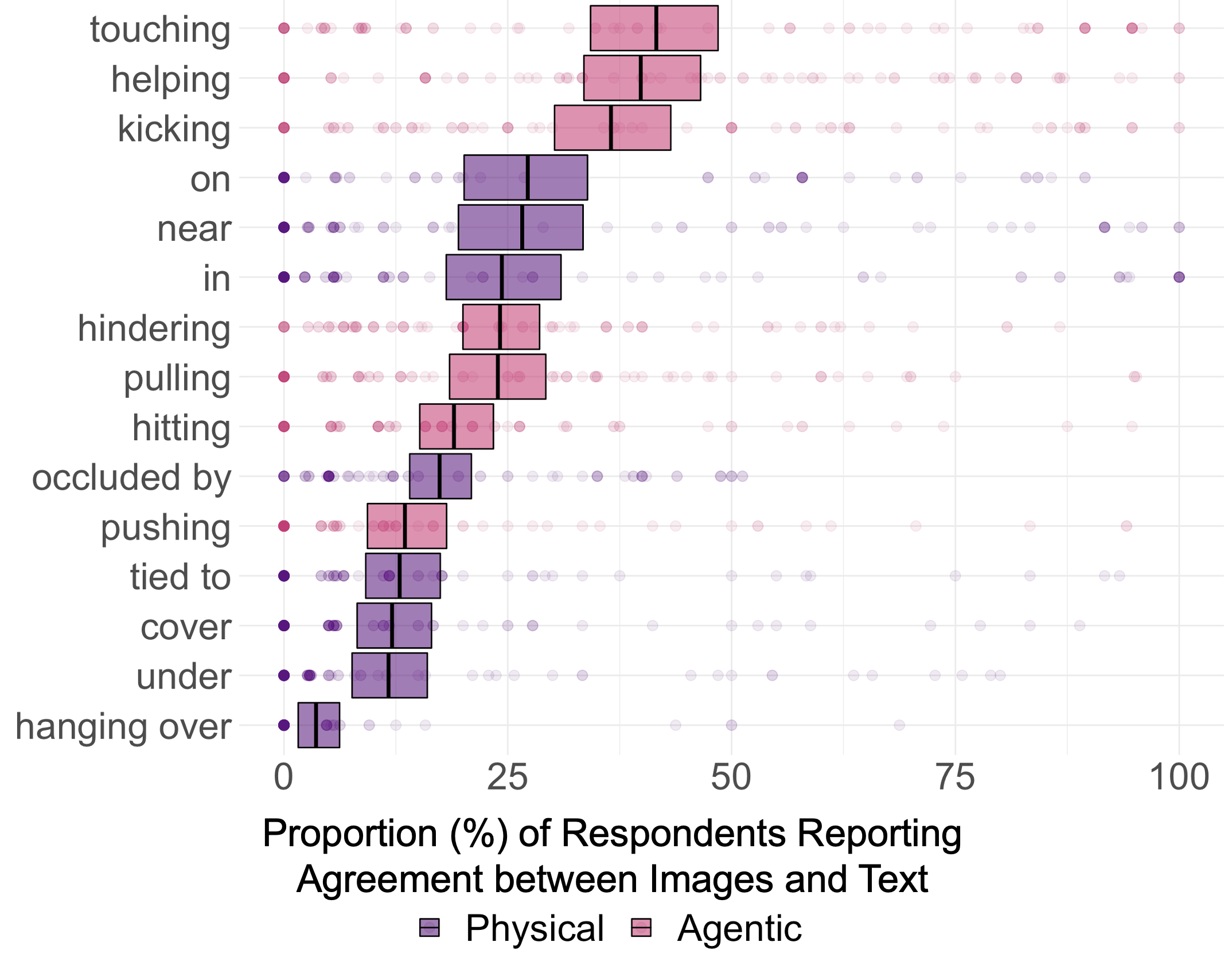}
\captionsetup{justification=justified,margin=0.5cm}
\caption{The proportion of participants reporting agreement between image and prompt, by the specific relation being tested. Points are the means of individual images, averaged across participants. There is a large range of reported agreement between image and text, though no relation entails average agreement significantly greater than 40\%.}
\label{figure:relation-scores}
\end{figure*}

Decomposing the broad categories of physical and agentic into constituent relations, we observe a range of human agreement scores, as shown in Figure  \ref{figure:relation-scores}. While it is difficult to say what criterion establishes whether DALL-E 2 'understands' a given relation, here we report comparisons to 3 thresholds: 0\%, 25\%, and 50\% perceived agreement, averaged across participants. Holm-corrected, one-sample significance tests for each relation suggest all 15 relations have participant agreement significantly above 0\% at $\alpha = 0.95$ ($p_{Holm} < 0.05)$). However, only 3 relations entail agreement significantly above 25\% (touching, helping, and kicking), and no relations entail agreement above 50\%. This remains true even without correction for multiple comparisons. 

\begin{figure*}[h!]
\centering
\includegraphics[width=0.75\linewidth]{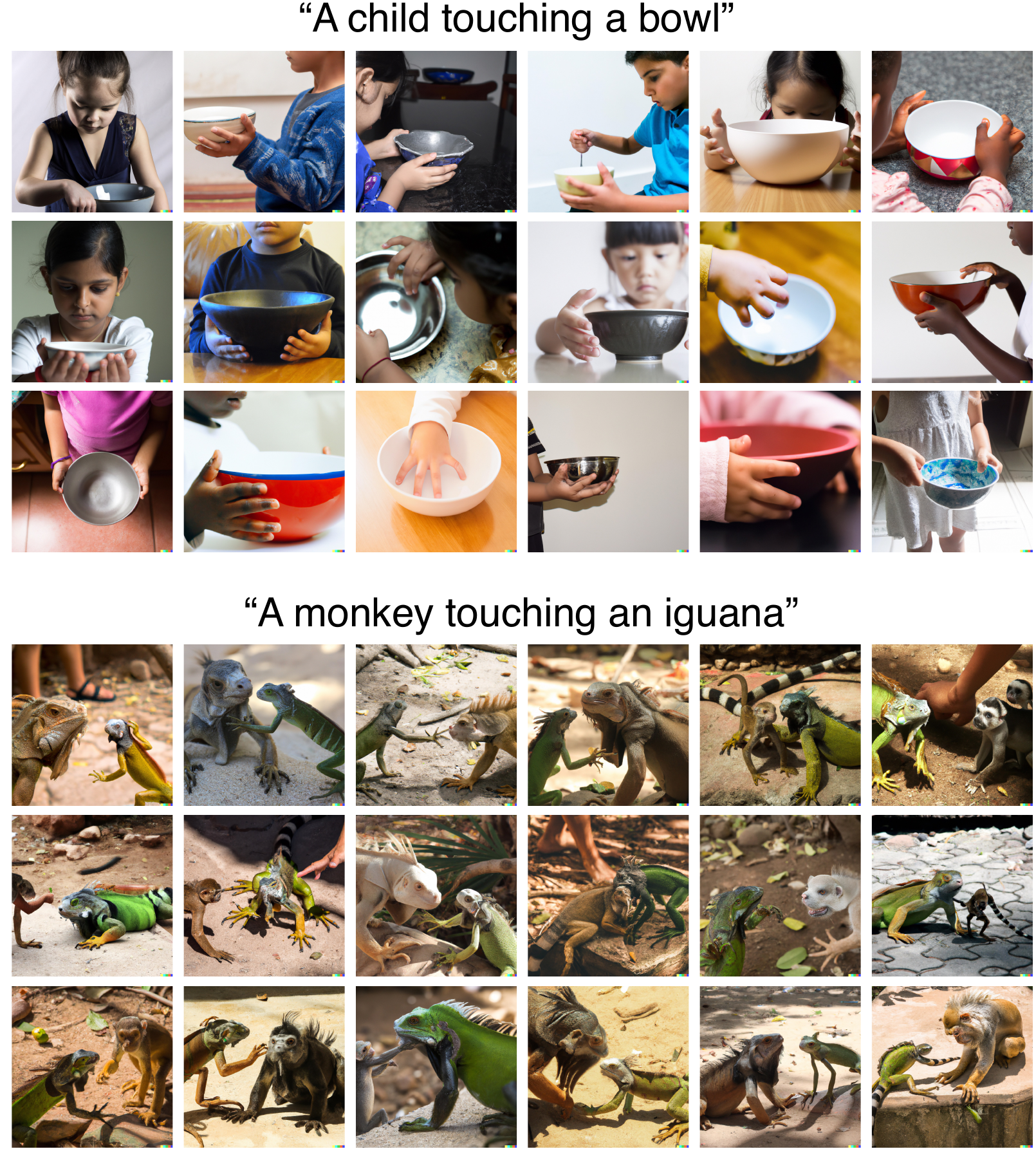}
\captionsetup{justification=justified,margin=0.5cm}
\caption{Grids for two example prompts that probed the \textit{touching} relation. While the average agreement was 42\%, the underyling distribution of prompt responses was effectively bimodal, with e.g. the prompt `a child touching a bowl' generating high agreement (87\%), and `a monkey touching an iguana' generating low agreement (11\%).}
\label{figure:best-worst}
\end{figure*}

Considering the results qualitatively, we note that even a (relatively) high average agreement may not indicate relational understanding, but rather an influence of the training set. For example, the `touching' relation generated maximal average agreement (at a mean of 42\%  [34.3,  49.6] across 90 images), but with varied, bimodal success at the level of individual prompts. For example, the prompt `child touching a bowl' generated 87\% [80.1, 93] agreement on average, while `a monkey touching an iguana' generated 11\% [5.3, 19.7] agreement on average (see Figure \ref{figure:best-worst}). It may be then that the combination of `child' and `bowl' is likely to generate images of a child touching a bowl simply given the training data. We consider this point further in the discussion. 

\begin{figure*}[h!]
\centering
\includegraphics[width=0.75\linewidth]{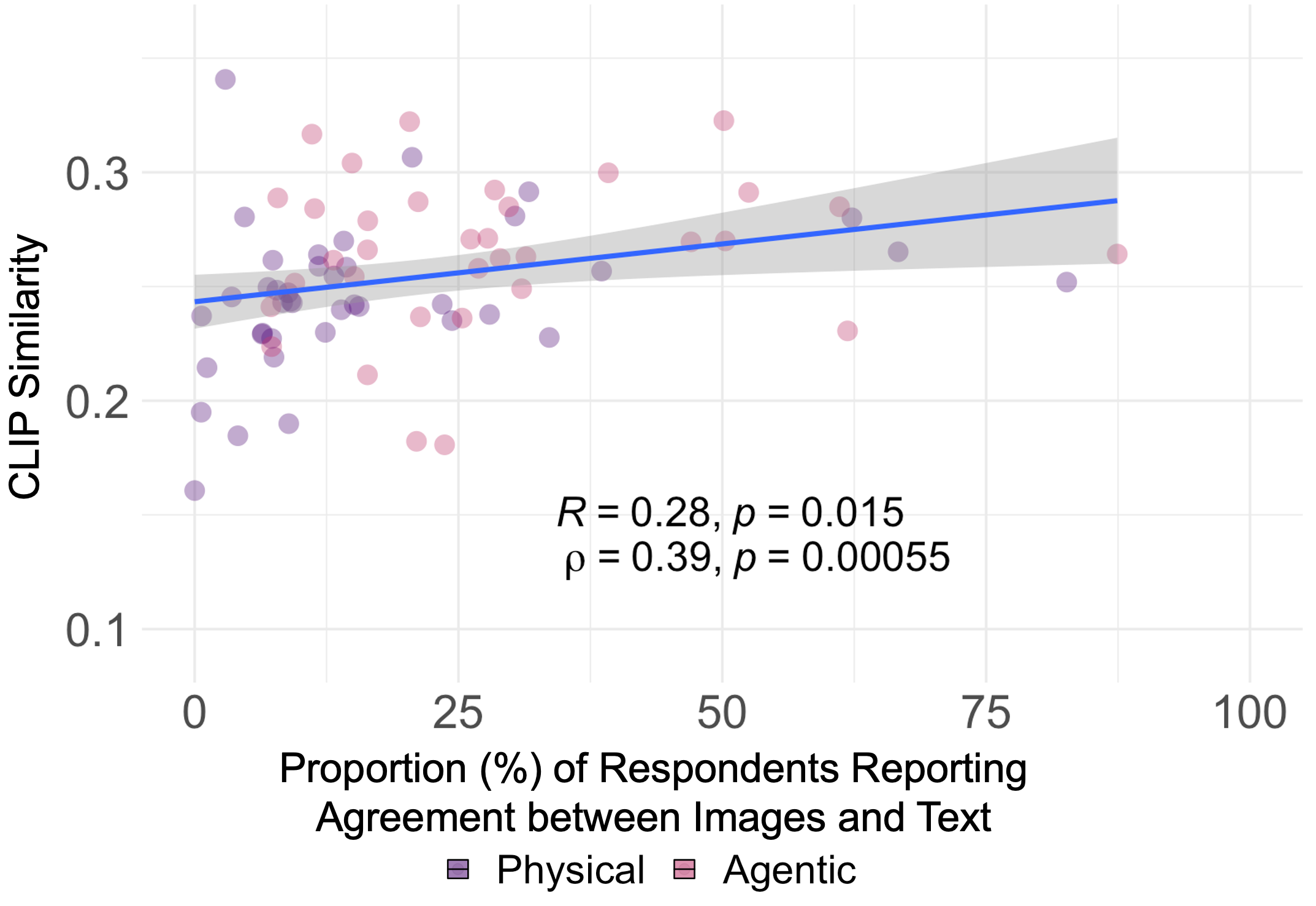}
\captionsetup{justification=justified,margin=0.5cm}
\caption{Relationship between CLIP (ViT-L/14) similarity scores and human agreement scores, averaged over images and participants. Each point is 1 / 75 prompts.}
\label{figure:clip-versus-human}
\end{figure*}

While there are many factors that influence the quality of DALL-E 2's generated outputs, one particular parameter of interest is the CLIP score of the generated images: That is, the similarity (as determined by CLIP) between the generated image, and the text prompt used to generate that image \citep{radford2021learning}. Intuitively, this is one of the parameters most responsible for the match between the target linguistic concept (in this case, a relation) and its depiction, but it's not necessarily a given that CLIP accounts for relations specifically. To examine the relationship between CLIP similarity and human perception, we used OpenAI's open-source ViT-L/14 model to calculate the similarity score between each image in our image set and their associated prompts. We then averaged the CLIP scores across the 18 images generated from each prompt, and correlated this average with the average perceived agreement provided by the human respondents. We found a moderate relationship between the two: $\hat{\rho}_{Spearman} = 0.39 \:[0.17,0.57], p = 5.5e-4 $ (and see Figure  \ref{figure:clip-versus-human}), suggesting CLIP is at least partially sensitive to the kinds of relations we've tested. 

\newpage
To assess more finely the combined influence of broad relation type ('agentic' or 'physical') and CLIP scores on the human-perceived match between text and image, we used two Bayesian multilevel (mixed-effects) models: a zero-inflated binomial model calculated directly over the participant-level choice data (with additive effects for relation type and CLIP score, plus random intercepts for subject and the order of image presentation [0-18]), and a zero-one-inflated beta model calculated over the average scores per image (again with additive effects for relation type and CLIP score, but with a random intercept for the order of image presentation alone). We use zero-inflated models in both cases, given the outsize quantity of images that participants labeled as not matching the target prompt. Controlling in both cases for variance injected by factors outside the study design (i.e. random effects), these models suggest small-to-midsize significant effects of both relation type and CLIP score on the probability of human respondents designating a target image as matching its prompt. Results from these regressions are summarized in Table  \ref{results-regression}.

\begin{table*}[h!]
\centering
\includegraphics[width=0.85\linewidth]{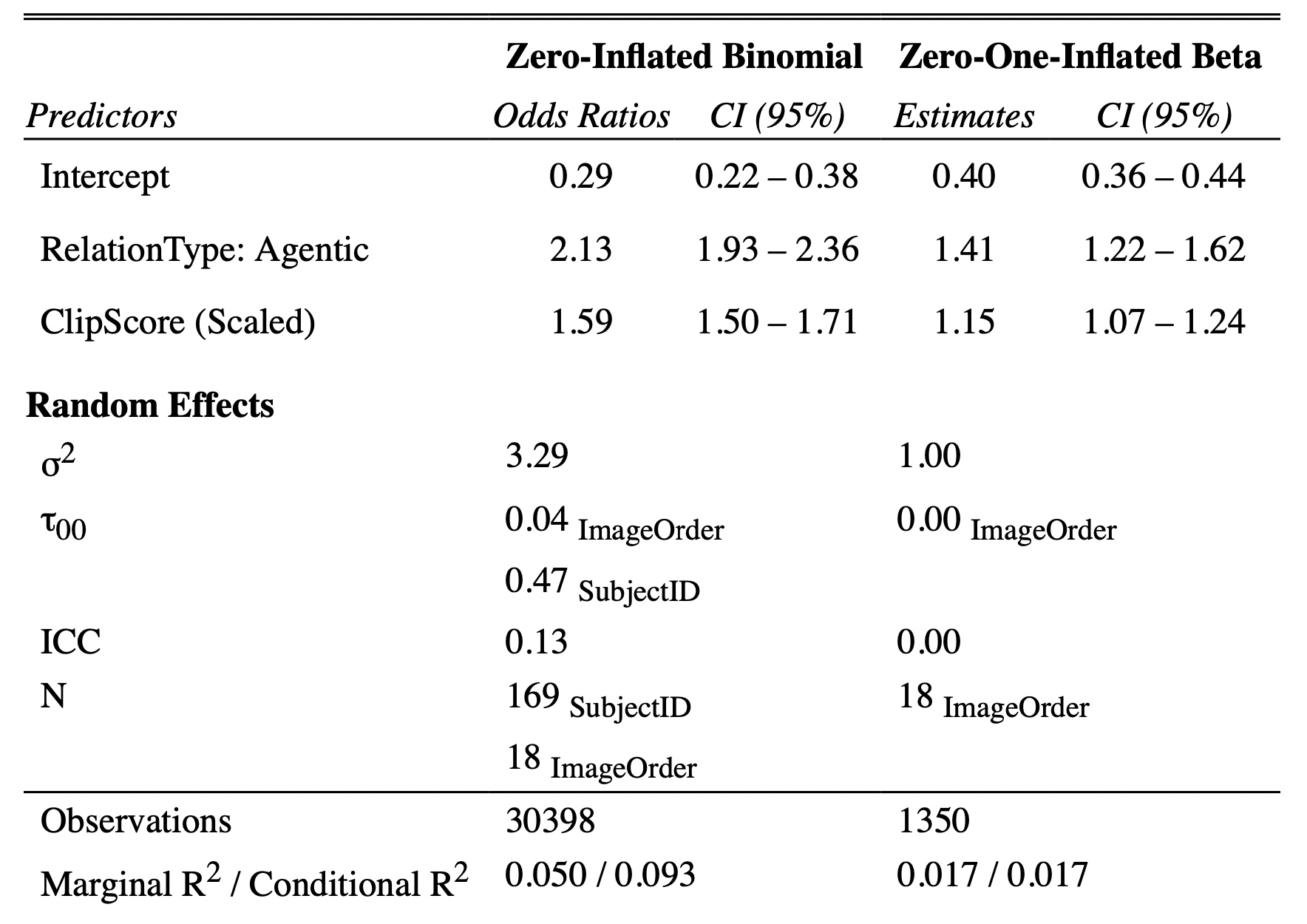}
\captionsetup{justification=justified,margin=0.5cm}
\caption{Results of two mixed effects regressions of relation type and CLIP score on human agreement, either at the individual subject level (zero-inflated binomial) or the image level (zero-one-inflated-beta).}
\label{results-regression}
\end{table*}

\section{Discussion}

DALL-E 2 and its two constituent components (latent diffusion models and CLIP) represent a significant advance in machine learning, and such models may well spur new directions in the creative visual arts. However, our current experiment and analysis suggests that DALL-E 2 suffers from a significant lack of common sense reasoning in the form of relational understanding. 

Relational understanding is a fundamental component of human intelligence, which manifests early in development \citep{spelke1994initial}, and is computed quickly and automatically in perception \citep{hafri2021perception}. DALL-E 2's difficulty with even basic spatial relations (such as \textit{in, on, under}) suggests that whatever it has learned, it has not yet learned the kinds of representations that allow humans to so flexibly and robustly structure the world. A direct interpretation of this difficulty is that systems like DALL-E 2 do not yet have relational compositionality. 

The notion that systems like DALL-E 2 do not have compositionality may come as a surprise to anyone that has seen DALL-E 2's strikingly reasonable responses to prompts like `a cartoon of a baby daikon radish in a tutu walking a poodle'. Prompts such as these often generate a sensible approximation of a compositional concept, with all parts of the prompts present, and present in the right places. Compositionality, however, is not only the ability to glue things together -- even things you may never have observed together before. Compositionality requires an understanding of the \textit{rules} that bind things together. Relations are such rules. 

\begin{figure*}[h!]
\centering
\includegraphics[width=0.85\linewidth]{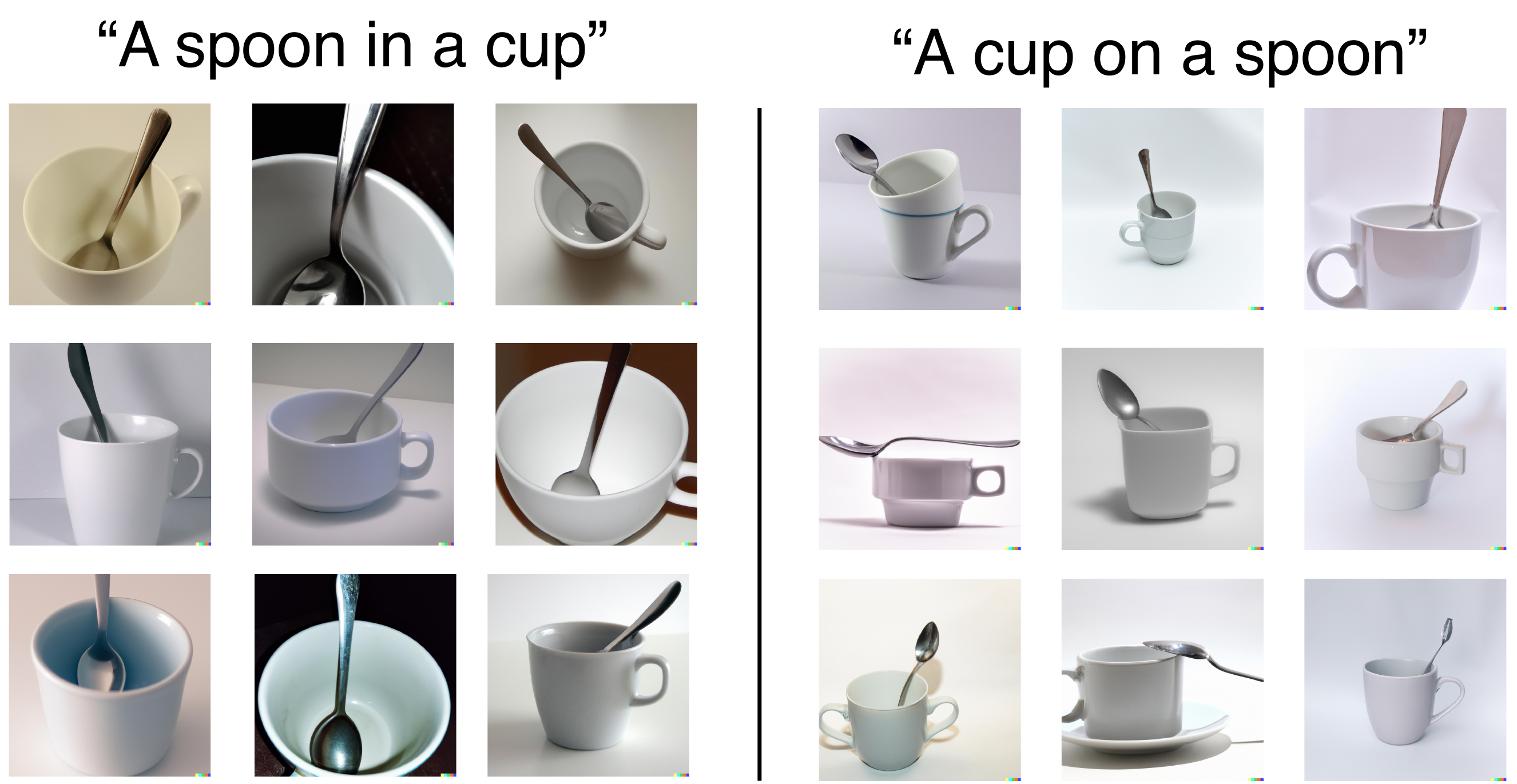}
\caption{Illustrative example, images generated given `a spoon in a cup' and `a cup on a spoon'. Examining just the left images may lead to the conclusion that Dall-E 2 captures the \textit{in} relation, but the right images suggest this is simply an effect of training images that involve \textit{spoon} and {cup}.}
\label{figure:cup-spoon}
\end{figure*}

To the extent that DALL-E 2 is only able to generate relations some of the time is the extent to which DALL-E 2 is actively \textit{not} compositional. These failure cases are important, because they tell us something about the way DALL-E 2 is getting things \textit{right}. The fact that DALL-E 2 seems able to easily generate `a spoon in a cup', but not `a cup on a spoon' (see Figure  \ref{figure:cup-spoon}), means that even when it is getting `a spoon in a cup' right this is likely due to a great deal of prior exposure to images of spoons in cups, rather than an understanding of `in' or `on' -- precisely the kinds of syntactic rules that define compositionality. Real compositionality should be invariant at the level of the relation, which is to say that ambiguity in meaning should come from the semantic elements involved in the relation, and not from the relation itself \citep{pelletier1994principle,pelletier2016semantic}.

In addition to effects of training data on apparent successes, it is possible that DALL-E 2's slightly better performance with more abstract relations like 'helping' is due to visual ambiguities, and the interpretive steps that people take on top of a given image. That is, when seeing an image of a robot touching another robot and the prompt `a robot helping a robot', people may be thinking `Well, I guess this \textit{could} be helping, if...'. This is a tentative suggestion, but it could be tested empirically by showing people images generated through prompts like `helping' but without labeling, and having them either freely describe the image, or giving people a forced choice among several relations.

Even with the occasional ambiguity, the current quantitative gap between what DALL-E 2 produces and what people accept as a reasonable depiction of very simple relations is enough to suggest a \textit{qualitative} gap between what DALL-E 2 has learned, and what even infants seem already to know. This gap is especially striking given DALL-E 2's staggering diet of image content. 

There are many potential reasons for Dall-E 2's current lack of relational understanding, and they range from the minutia of technical implementation, to larger disjuncts between the computational principles underlying human intelligence and those underlying many current artificial intelligence systems. One such disjunct is the way in which `place' is explicitly coded for in both the generative image and text models that constitute text-guided image generation algorithms like DALL-E 2. Perhaps the only explicit encoding of relational order in such models is to be found in the positional embeddings of the text transformer in CLIP -- effectively an auxiliary input that might easily be outweighed by the dozen or so nonlinear attention heads between them and the model's final outputs. This design choice is a marked difference from earlier iterations of natural language processing algorithms that provide syntactic parse trees in conjunction with the tokens corresponding to individual morphemes and words \citep{tai2015improved}. At the level of images, there is an incompatibility between many modern machine vision algorithms -- often designed \textit{explicitly} to mimic the primate ventral visual stream -- and the explicit representation of relations (spatial and otherwise) in the primate dorsal stream \citep{summerfield2020structure}. Text-guided image generation algorithms might well benefit from mimicking algorithms in robotics \citep[e.g. CLIPort][]{shridhar2021cliport}, which combine CLIP's semantic flexibility with spatial transformers to model object identities and affordances simultaneously. 

Another plausible upgrade that may boost model performance on relations are architectural adjustments that allow for multiplicative effects in a single layer of computation \citep{steinberg2022associative}. These kinds of adjustments are inspired by biological perceptual systems, including the dorsal stream, that contain mixed selectivity neurons and lateral sub-circuits that facilitate the representation of interactions at multiple levels of the information-processing hierarchy \citep{silver2010neuronal, fusi2016neurons}. 

DALL-E 2 and other current image generation models are things of wonder, but they also leave us wondering what exactly they have learned, and how they fit into the larger search for artificial intelligence. DALL-E 2 has seemingly done what many models before it have failed to do, and bound the abstractions of natural language to clear points of perceptual reference. But that binding so far remains far more tenuous than the binding that defines the clear referents of standard human communication. The case of relational understanding provides a clear target for making an already meaningful advancement in artificial intelligence even closer to human meaning. 

\subsection*{Acknowledgments}

We thank OpenAI for providing access to DALL-E-2 engine, and Jiayi Wang for her help in creating the stimuli. This work was supported by the Center for Brains, Minds and Machines (CBMM), funded by NSF STC award CCF1231216.

\bibliographystyle{vancouver}
\bibliography{references}

\appendix

\counterwithin{figure}{section}
\section{Appendix}

\begin{figure*}[h!]
\centering
\includegraphics[width=0.85\linewidth]{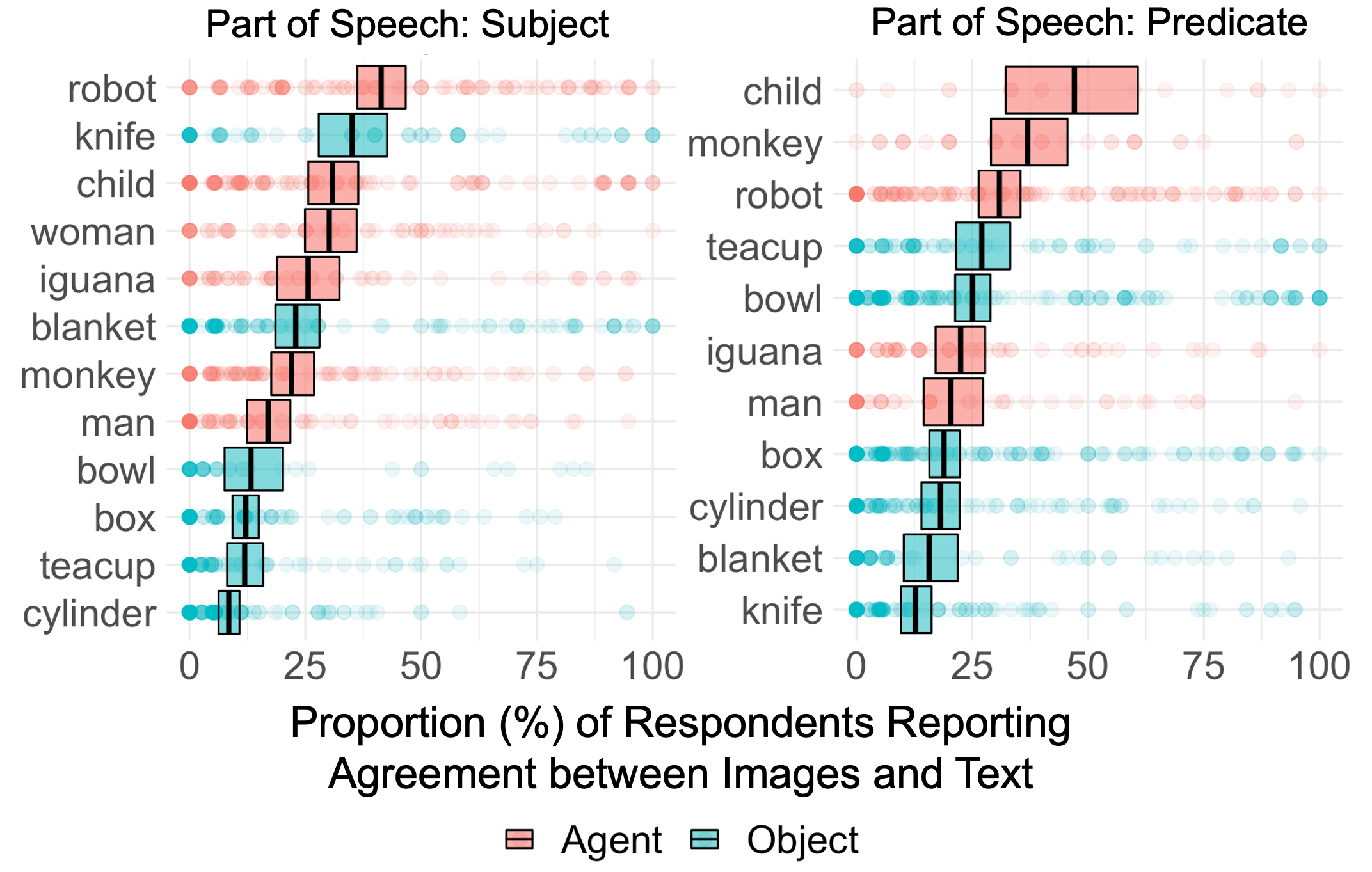}
\captionsetup{justification=justified,margin=0.5cm}
\caption{The proportion of respondents reporting agreement between image and prompt, broken down by each entity's part of speech.}
\label{figure:pos-scores}
\end{figure*}

\end{document}